\documentclass[letterpaper, 10 pt, conference]{ieeeconf}  
\usepackage{amsmath,amssymb,amsfonts}
\usepackage{algorithmic}
\usepackage{graphicx}
\usepackage{textcomp}
\usepackage{xcolor}
\usepackage{subcaption}
\usepackage{multirow}
\usepackage{cuted}
\usepackage{booktabs}
\usepackage{multirow}
\usepackage{threeparttable}
\usepackage{tabularx}
\usepackage{soul}
\usepackage[autostyle]{csquotes}
\usepackage{balance}
\usepackage{hyperref}

\IEEEoverridecommandlockouts                              

\title{\LARGE \bf
  What Is My Robot Thinking? Design Considerations for Transparent and Trustworthy Shared Autonomy
}

\author{Atharv Belsare$^{1*}$, Zohre Karimi$^{1,2}$, Connor Mattson$^{1,2}$, Rushiil Nakka$^{1}$ and Daniel S. Brown$^{1,2}$%
\thanks{*Corresponding author: atharv.belsare$@$utah.edu}
\thanks{$^{1}$Kahlert School of Computing, University of Utah, Salt Lake City, UT, USA.}%
\thanks{$^{2}$Robotics Center, University of Utah, Salt Lake City, UT, USA.}%
}

\begin{document}
\maketitle
\thispagestyle{empty}
\pagestyle{empty}

\begin{abstract}
Assistive robots operating under shared autonomy must balance user control with autonomous assistance. Because robot actions depend on internal intent inference that is not directly observable, mismatches between inferred and intended goals can undermine coordination and trust. We investigate how interface-level transparency--feedback modality (visual vs.\ auditory) and information richness (sparse vs.\ rich)--shapes interaction in a vision-based shared autonomy system. In a user study ($N=25$) across two assistive manipulation tasks, we evaluate how these designs influence coordination and trust.
Providing feedback significantly improves intent alignment and reduces corrective intervention, indicating that making the inferred goal legible accelerates convergence in shared control. Participants preferred visual over auditory feedback, while preferences for sparse versus rich information depended on task complexity. We also found that revealing the full belief distribution did not consistently improve alignment or trust. Together, these findings indicate that effective transparency enhances coordination primarily through goal legibility, while trust depends on task-appropriate information exposure rather than maximal disclosure. Based on these results, we outline guidelines for designing transparent shared autonomy systems.
Code and videos are available at \url{https://sites.google.com/view/design-t2-sa/home}.
\end{abstract}

\begin{figure}[!ht]
  \centering
  \begin{subfigure}{\columnwidth}
    \centering
    \includegraphics[width=0.85\columnwidth]{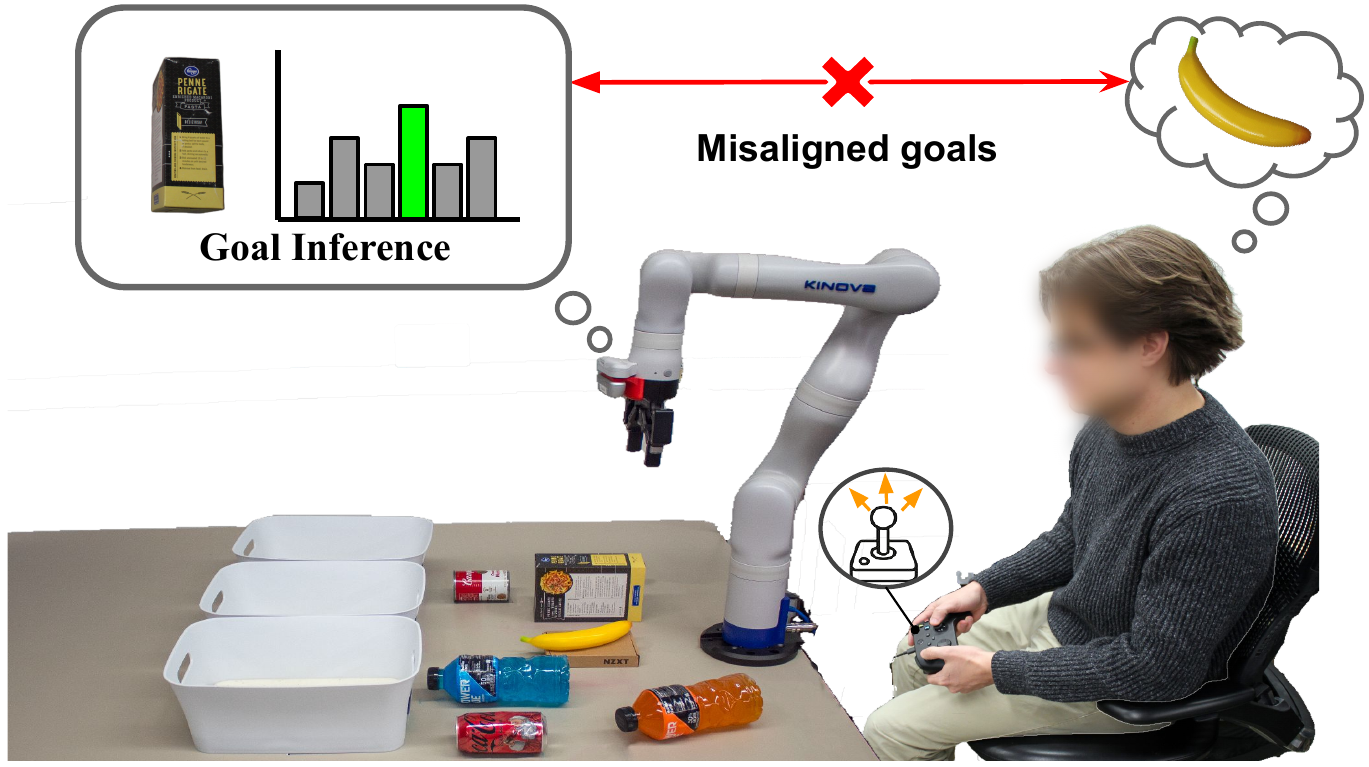}
  \end{subfigure}
  \begin{subfigure}{\columnwidth}
    \centering
    \includegraphics[width=0.85\columnwidth]{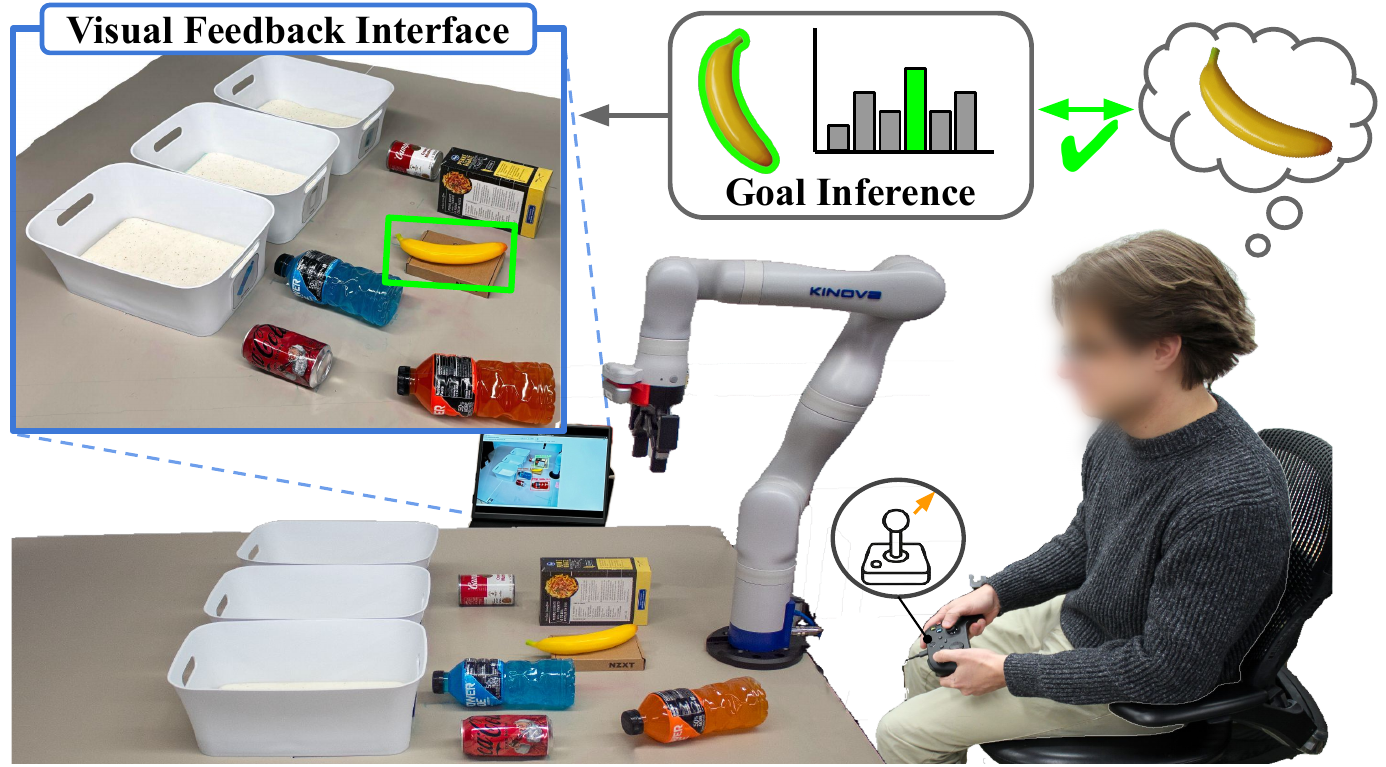}
  \end{subfigure}
  \begin{subfigure}{\columnwidth}
    \centering
    \includegraphics[width=0.85\columnwidth]{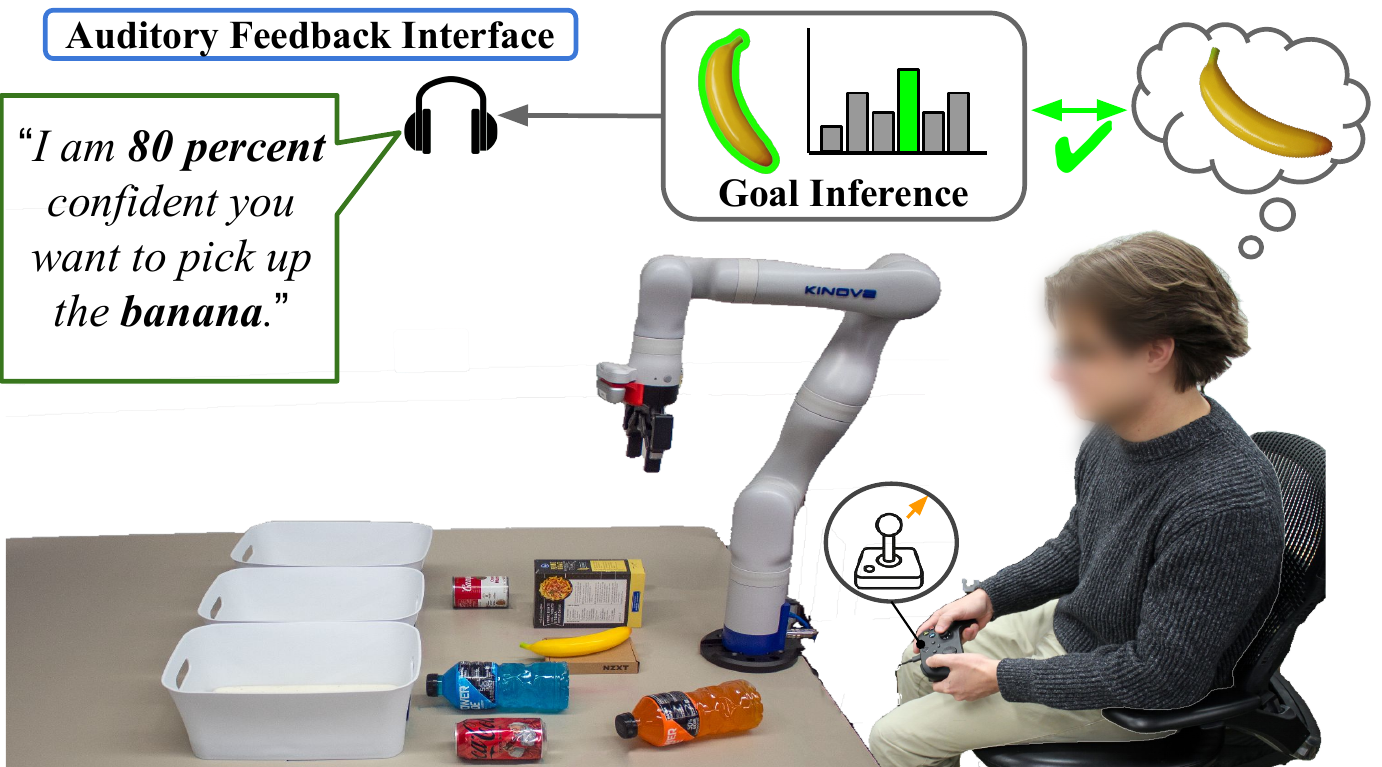}
  \end{subfigure}
  \caption{(\textbf{Top}) The robot infers a goal from the human’s joystick input but does not communicate this inference. This can lead to misaligned goals and requires the human to guess whether the robot needs a correction. (\textbf{Middle}) With a visual feedback interface, the robot displays its current inferred intent, allowing the human to see what the robot believes the target to be and adjust their input accordingly. (\textbf{Bottom}) With auditory feedback, the robot verbally communicates its inferred intent, similarly supporting intent alignment. 
  }
  \label{fig:teaser}
\end{figure}

\section{Introduction}
Assistive robotics is a central area of research in human--robot interaction (HRI), aimed at supporting users in performing everyday tasks safely and independently. Such systems are increasingly deployed in domains ranging from physically demanding or complex manipulation tasks performed by able-bodied users~\cite{losey2022learning, mitzner2018closing} to individuals with motor impairments~\cite{taylor2018americans, argall2018autonomy}. Examples include shared-control manipulation systems that infer user intent and enable human–robot mutual adaptation during collaborative tasks~\cite{8593766, nikolaidis2017human}, as well as rehabilitation exoskeletons for post-stroke recovery and wheelchair-mounted manipulators that enable users to interact with their environment~\cite{beckerle2017human, carlson2012collaborative}. In these settings, users rely on robotic assistance, underscoring the importance of systems that behave in a predictable, interpretable, and trustworthy manner.

Many assistive robotic platforms are high degree-of-freedom systems controlled through low-dimensional interfaces such as joysticks, which can be cognitively demanding~\cite{meeker2018intuitive, jeon2020shared, belsare2025toward}. While full autonomy can reduce this burden, it may also diminish user agency and undermine trust~\cite{lee2004trust, onnasch2014human}. Shared autonomy addresses this trade-off by combining human input with autonomous assistance, typically through linear blending of user control and robot actions inferred from predicted user intent~\cite{belsare2025toward, dragan2012formalizing, udupa2023shared}. By allowing users to guide high-level intent while delegating low-level control, shared autonomy reduces workload while preserving flexibility. However, while trust and interpretability have been widely studied in automation and human--robot interaction more broadly~\cite{lee2004trust, onnasch2014human}, their role as explicit interface design variables in continuous shared autonomy remains underexplored.

In many shared autonomy systems, transparency is assumed to emerge implicitly through observable robot behavior, such as motion trajectories or goal-directed actions. However, when assistance depends on online intent inference, motion alone may not reveal the robot's internal inference state (see Fig.~\ref{fig:teaser}). While several shared autonomy works incorporate explicit feedback mechanisms, such as visual overlays or verbal cues, to communicate inferred goals during control~\cite{rosenthal2016verbalization, kim2006should, lakhmani2016proposed, alonso2018system}, these mechanisms are typically evaluated individually rather than systematically compared under identical autonomy policies. As a result, it remains unclear how feedback modality and information richness independently influence coordination and trust in shared autonomy.

In this work, we address this gap by experimentally evaluating how feedback interfaces influence transparency and trust in shared autonomy. We study visual and auditory feedback modalities, each presented in sparse and rich forms, within a vision-based shared autonomy system that infers user intent from egocentric visual observations~\cite{belsare2025toward}. By holding the shared autonomy controller fixed and varying only how intent information is presented, we isolate the effects of interface design on user interaction. We conduct a user study with 25 participants across two assistive manipulation tasks differing in complexity. Our results show that providing feedback significantly improves user--robot intent alignment and reduces the need for corrective interventions. However, we find that revealing the robot’s full belief distribution does not necessarily yield additional improvements in coordination or trust, underscoring the need for transparent designs that adapt to task demands and user needs.

Our main contributions are: 
(1) We develop and integrate a suite of feedback interfaces within a vision-based shared autonomy system, varying modality (visual vs. auditory) and richness (sparse vs. rich). 
(2) We evaluate these interfaces in a user study with 25 participants across two assistive manipulation tasks. 
(3) We outline design guidelines for transparent shared autonomy systems, showing that transparency improves coordination by enabling faster convergence to a stable shared intent, while trust depends on task-appropriate information rather than full belief disclosure.

\begin{figure*}
  \centering
  \includegraphics[width=\textwidth]{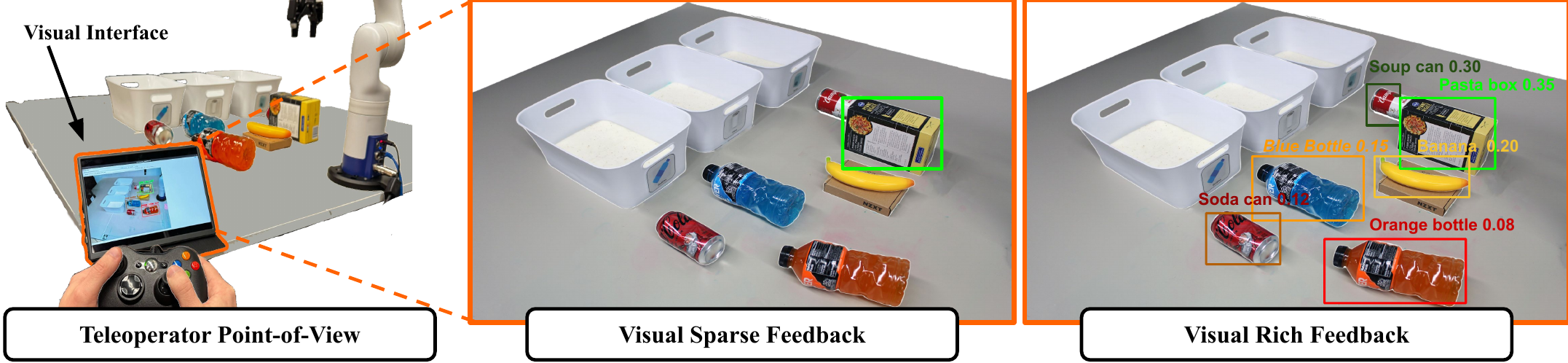}
  \caption{\textbf{Visual Feedback Interfaces.} 
  (\textbf{Left}) The teleoperator’s point-of-view, showing the scene camera feed on a display alongside the physical workspace.
  (\textbf{Middle}) \textit{Visual Sparse (VS)}: A single bounding box highlights the object with the highest inferred intent confidence, indicating the robot’s current predicted target. 
  (\textbf{Right})  \textit{Visual Rich (VR)}: All candidate intents are displayed with bounding boxes, object labels, and confidence scores representing the robot’s internal distribution over possible intents; bounding box color is drawn from a red-green spectrum to encode confidence (green = high, red = low).
  }
  \label{fig:interfaces}
\end{figure*}



\section{Related Work}
Transparency has been conceptualized in multiple ways across human–robot interaction. Behavior-centric perspectives relate transparency to motion legibility and predictability, which support human inference of intent from robot trajectories~\cite{dragan2013legibility, dragan2015effects}. The Situation Awareness–based Transparency framework organizes transparency across perception, comprehension, and projection levels~\cite{chen2014situation}, while surveys describe dimensions such as observability and explainability in robotic systems~\cite{alonso2018system}. These perspectives clarify what transparency entails conceptually but provide limited empirical guidance on how transparency should be instantiated as a real-time interface variable during shared control~\cite{theodorou2017designing, wortham2017robot, schott2023literature}. In this work, we study transparency through feedback modality and information richness and evaluate their effects during continuous shared autonomy.

Within shared autonomy, several systems communicate internal inference during control. Hoegerman et al.~\cite{hoegerman2024aligninglearningcommunicationshared} expose inferred belief to users but couple transparency with modified learning updates. Mullen et al.~\cite{9536385} compare augmented reality and haptic feedback for communicating inferred goals, yet evaluate these within a specific multimodal architecture. Brooks and Szafir~\cite{brooks2020visualizationintendedassistanceacceptance} visualize intended goals and trajectories to influence acceptance of assistance but focus exclusively on visual feedback. Zolotas and Demiris~\cite{zolotas2019towards, 8594002} evaluate augmented reality visualizations of shared-control state but do not compare across other modalities. Zhang et al.~\cite{9560991} introduce haptic feedback to improve agreement in teleoperation but study a single modality without varying levels of information detail.
Across these systems, transparency is evaluated as part of a specific controller and feedback design rather than systematically varied as an independent interface factor. As a result, prior work does not isolate the independent effects of feedback modality and information richness under fixed shared autonomy policies. In contrast, we vary only how the robot’s internal state is presented, through modality and information richness, while keeping inference and blending policies fixed.

Assistive shared control introduces additional constraints compared to systems that provide explanations outside the control loop, as users must interpret autonomy while continuously contributing control input~\cite{dragan2012formalizing, carlson2012collaborative}. Blending strategies and interface design influence predictability and user workload~\cite{ezeh2017probabilistic}. Several systems rely on augmented reality or specialized haptic devices to externalize autonomy state~\cite{zolotas2019towards, 9560991}. To our knowledge, no prior shared autonomy systems have isolated and compared lightweight visual and auditory feedback under fixed blending policies. We therefore evaluate screen-based visual and auditory feedback within a shared autonomy framework in this work.

The degree of information revealed during shared control also remains underexplored. While richer belief or trajectory exposure has been associated with improved usability and predictability in individual systems~\cite{9536385, brooks2020visualizationintendedassistanceacceptance}, these studies do not isolate information richness as an independent factor under fixed shared autonomy policies. It remains unclear whether increasing belief information improves alignment and trust when autonomy is held constant. We study this by varying information richness across interface modalities.

Motivated by these limitations, we treat transparency as an interface design variable, defined by modality and information richness, and evaluate its independent effects under a fixed shared autonomy controller.

\section{Preliminaries}
\subsection{Transparency in Blended Shared Autonomy}
We consider a shared autonomy approach where user teleoperation input is combined with an autonomous policy to produce a robot action~\cite{dragan2012formalizing, belsare2025toward}. Let $X$ denote the robot state space and $U$ the action space, where $u_h, u_r \in U$ are the human and robot actions. A finite set of candidate goals or intents $G$ is inferred from scene observations. At time $t$, the system maintains a belief $b_t$ over $G$ and selects the most likely goal $\hat{g}_t \in G$ to compute the autonomous action $u_r(t)$.
Control is generated via linear blending:
\begin{equation} 
u(t) = (1 - \alpha)\, u_h(t) + \alpha\, u_r(t),
\end{equation}
where $\alpha \in [0,1]$ increases with confidence in $\hat{g}_t$~\cite{dragan2012formalizing}. Because $u_r(t)$ depends on $\hat{g}_t$, mismatches may arise between the user’s intended goal and the system’s estimate. Although the inference is deterministic, the robot's belief distribution $b_t$ over human intent is usually not observable to the user, which can create uncertainty about the robot’s interpretation.  

\subsection{Transparency Interface as a Design Variable}
Our shared autonomy system maintains $\hat{g}_t$ and $b_t$ at each time step. The transparency interface specifies how this internal inference state is communicated to the user through feedback~\cite{chen2014situation, alonso2018system}. We model the displayed feedback as:
\begin{equation} \label{z_t}
z_t = \phi(m, r; \hat{g}_t, b_t),
\end{equation}
where $z_t$ is the feedback signal, $\phi$ maps $(\hat{g}_t, b_t)$ to user-visible or user-audible output, $m \in \{\text{visual}, \text{auditory}\}$ denotes modality, and $r \in \{\text{sparse}, \text{rich}\}$ denotes information richness. Sparse feedback communicates only $\hat{g}_t$, while rich feedback additionally exposes belief information derived from $b_t$.
We define \emph{goal legibility} as the extent to which the robot’s current predicted goal $\hat{g}_t$ can be clearly identified by the user during interaction. Unlike motion-based legibility, we treat it as an interface-level property determined by how explicitly $\hat{g}_t$ is conveyed through $z_t$. By varying $(m,r)$ while keeping the controller fixed, we isolate how interface-level transparency influences coordination and trust.

\section{Interface Design} \label{sec:interface-design}
Motivated by the limited comparison of feedback modalities in shared autonomy, and by evidence that visual and auditory feedback are effective for intent communication in other domains, we investigate these modalities as transparency mechanisms in our system. We first describe the visual interfaces (see Fig.~\ref{fig:interfaces}), followed by the auditory interfaces. Each interface was implemented as an instance of $\phi$ in Eq.~\ref{z_t}, producing feedback $z_t$ that varies in modality $m$ and information richness $r$.
\subsection{Visual Interfaces}
\subsubsection{Visual Sparse}
In this interface, the system uses a single bounding box to highlight the object that the robot currently infers as the user’s intended target. After a successful grasp, the interface highlights the most likely placement location inferred, enabling the user to anticipate the robot’s next action. This design emphasizes clarity and reduces cognitive overhead by presenting only the robot’s top prediction, aligning with transparency frameworks that emphasize perception-level cues~\cite{chen2014situation}.
\subsubsection{Visual Rich}
In this interface, the system uses bounding boxes to highlight all candidate objects and overlays belief information for each. For every object, it displays a text label and confidence score, with bounding-box color encoding confidence (green = high, red = low). After a successful grasp, the interface highlights candidate placement locations and displays their associated confidence values, allowing users to view both the current prediction and alternative options. This design exposes the robot’s belief distribution over all candidate intents.

\subsection{Auditory Interfaces}\label{subsec:auditory-interfaces}
\subsubsection{Auditory Sparse}
In this interface, the system provides spoken feedback only when the robot updates its predicted intent. For example, if the prediction switches from a \textit{bottle} to a \textit{cup}, the system verbalizes the new object label. In this condition, only the currently inferred label is spoken. After a successful grasp, the system announces the most likely placement location inferred by the robot, allowing the user to anticipate where the robot can place the item. This design surfaces key changes in the robot's inference while minimizing distraction~\cite{rosenthal2016verbalization}.

\subsubsection{Auditory Rich}
In this interface, the system verbalizes belief information for all candidate objects whose confidence exceeds a predefined threshold. For each object, it communicates the label and confidence score associated with the predicted action. For example, the robot might say: \textit{“I am 60 percent confident you want to grasp the bottle, and 40 percent confident you want to grasp the cup.”} After a successful grasp, the system similarly verbalizes confidence information for placement locations above the same threshold, allowing users to hear both the robot’s current prediction and its alternatives. This design makes alternate intents explicit, consistent with prior findings that explanations can support trust in uncertain contexts~\cite{hayes2017improving}.

\begin{figure}[!t]
  \centering
  \includegraphics[width=\columnwidth]{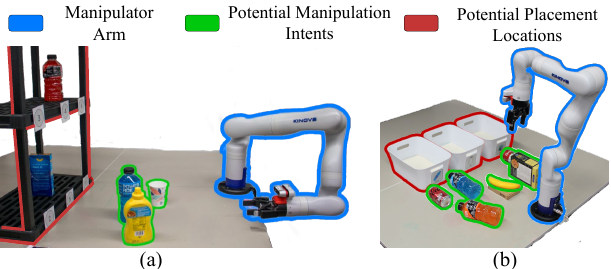}
  \caption{\textbf{User Study Tasks.}
    (a) Shelving task where users were asked to place objects onto designated shelves.
    (b) Sorting task where the robot helps the user sort multiple objects into their respective recycling bins (cardboard, plastic, metal).
  }
  \label{fig:experiment-setup}
\end{figure}

\section{Experiments}
\begin{figure*}[t]
  \centering
  \includegraphics[width=\textwidth]{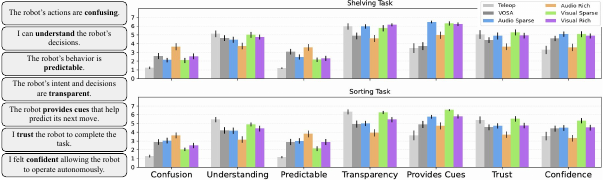}

  \caption{\textbf{Post Treatment Questionnaire and User Responses.} Mean post-trial 7-point Likert responses to the survey questions shown on the left, across interface conditions in the Shelving and Sorting tasks. Error bars denote standard error. Conditions include Teleop, VOSA (no feedback), and VOSA with visual or auditory feedback as described in Sec.~\ref{sec:interface-design}.}
  \label{fig:radar-plot}
\end{figure*}
We conducted a user study with 25 participants using a Kinova Gen3 7-DoF manipulator equipped with an Intel RealSense D435i depth camera mounted on the end-effector. An additional overhead RealSense D435i scene camera was installed to provide a global view for generating visual feedback.
Participants controlled the robot using an Xbox 360 controller, with the left joystick mapped to x–y planar motion and the right joystick to vertical (z-axis) motion.

\subsection{Implementation Details}
We implemented our interfaces on the recently proposed Vision-Only Shared Autonomy (VOSA) framework~\cite{belsare2025toward}. VOSA was selected for two key reasons. First, it enables real-time inference of user intent directly from egocentric visual observations, allowing continuous interaction. Second, because VOSA does not rely on task-specific training, predefined task models, or privileged state information, it is well suited for deployment in unstructured, real-world settings. This generality makes intent inference inherently uncertain and opaque to users, thereby motivating the need for explicit transparency mechanisms and making VOSA an ideal platform for this study.

All feedback interfaces were implemented without modifying VOSA's underlying intent inference or control logic. VOSA is a vision-based shared autonomy framework that performs online intent inference by estimating the user’s intent from joystick inputs and egocentric visual observations. At each timestep, VOSA estimates which object the user intends to grasp and provides (i) the object label, (ii) a confidence score, and (iii) the object’s location. These intent estimates were passed to the corresponding visual or auditory feedback interfaces, as described in Section~IV, such that all experimental conditions relied on identical underlying intent predictions and differed only in their presentation to the user.
For object detection, we fine-tuned YOLOv11~\cite{yolo11_ultralytics} on a custom dataset of task objects to ensure reliable detections. The same detector was used across all treatments to maintain identical inference. Auditory feedback was generated using the Piper text-to-speech library with fixed templates populated by the predicted label and confidence score. We used template-based generation to ensure consistent phrasing and low latency during real-time interaction. For the Auditory Rich condition, belief information was verbalized only for candidate intents whose confidence exceeded a threshold of 40 percent (tuned based on a pilot study), suppressing low-probability candidates and limiting excessive verbosity during real-time interaction.
Visual feedback was displayed on an external tablet, and auditory feedback was delivered via headphones.

\subsection{Baselines and Treatments}
Each task in the user study evaluated six treatments, grouped into no-feedback and feedback conditions. The two no-feedback treatments were Direct Teleoperation (\textbf{Teleop}), in which users controlled the robot using a joystick without autonomous assistance, intent inference, or feedback; and Vision-Only Shared Autonomy (\textbf{VOSA}), in which the system provided autonomous assistance but did not explicitly communicate its inferred intent.
The remaining four treatments incorporated explicit feedback on top of VOSA. These included two visual feedback conditions---Visual Sparse (\textbf{VS}) and Visual Rich (\textbf{VR})---and two auditory feedback conditions---Auditory Sparse (\textbf{AS}) and Auditory Rich (\textbf{AR}). Treatment order was randomized within each task, and participants were informed of the condition being evaluated.

\subsection{Hypotheses} \label{hypotheses}


We test the following hypotheses:
\textbf{H1}: Providing feedback will lead to improved alignment between the robot's intent and the users' true intent.
\textbf{H2}: Users will provide fewer corrective inputs when using feedback interfaces.
\textbf{H3}: Users will indicate increased understanding of the robot's goals with visual interfaces than with auditory interfaces.
\textbf{H4}: Users will indicate increased trust in the robot with visual interfaces than with auditory interfaces.
\textbf{H5}: Users will report higher trust in the robot system when provided with feedback compared to when no feedback is present.
\textbf{H6}: Users will prefer rich feedback interfaces over sparse feedback interfaces.
\textbf{H7}: Users will prefer visual feedback interfaces over auditory feedback interfaces.


\subsection{Task Description}
Prior to the main study, participants completed a practice session to familiarize themselves with the robot, control interface, and all experimental conditions. During this session, they experienced each of the six conditions---including Teleop, VOSA without feedback, and all four interface treatments---by performing a pick-and-place task. This allowed participants to gain exposure to the system's behavior under each condition and understand the type of information conveyed through the visual and auditory interfaces. Following the practice session, participants completed two tasks.

\subsubsection{Shelving}
This task was inspired by the shelving scenario introduced by Belsare et al.~\cite{belsare2025toward}. Participants were asked to assist in organizing groceries by using the robot to place two categories of items (bottles and condiments) onto a two-level shelf. All items were pre-arranged on the table at the start of the task, with sports drinks designated for the top shelf and mustard bottles for the bottom shelf (Fig.~\ref{fig:experiment-setup}a).

\subsubsection{Sorting (Recycling)}
At the start of the task, six objects were arranged in a cluttered workspace on the table. Participants collaborated with the robot to recycle three items in a predefined sequence: first, place an empty pasta box in the cardboard recycling bin; second, place a plastic bottle in the plastic recycling bin; and third, place an empty metal can in the metal recycling bin. Compared to the shelving task, this scenario was more complex due to the presence of multiple objects and increased clutter (Fig.~\ref{fig:experiment-setup}b).

\subsection{Study Protocol}
We recruited 25 participants (19 male, 6 female; average age $28.04 \pm 7.87$) from the university campus for an IRB-approved within-subjects user study. All participants provided informed consent. The study lasted approximately 75 minutes per participant, and received a \$25 compensation.
\subsection{Metrics Evaluated}
\textit{Objective Metrics:} We recorded the full robot state, joystick inputs, the predicted intent from the shared autonomy system throughout each trial. From these data, we extracted several task-relevant metrics aligned with our experimental hypotheses. For each task, the set of possible target objects and the high-level task structure were predefined as part of the experimental protocol, enabling all quantitative metrics to be computed automatically from system logs.

\textit{Subjective Metrics:} After each treatment, participants completed a 7-point Likert-scale questionnaire assessing their perceptions of the robot’s behavior. After each task, they selected their preferred treatment and could provide optional open-ended feedback. After completing both tasks, participants filled out a separate post-study questionnaire directly comparing interface modalities. The full set of per-treatment Likert questions and aggregated responses is shown in Fig.~\ref{fig:radar-plot}.



\section{Results} \label{sec:results}
\begin{table}[!t]
  \centering
    \caption{Participant interface preferences ($N=25$). Teleop and VOSA are no-feedback baselines; remaining conditions provide visual or auditory feedback. Percentages indicate overall interface preferences.}
  \label{tab:interface_preferences}
  \setlength{\tabcolsep}{10pt}
  \renewcommand{\arraystretch}{1.1}
  \begin{tabular}{lcc}
    \toprule
    \textbf{Interface} & \textbf{Shelving} & \textbf{Sorting} \\
    \midrule
    Direct Teleoperation (\textbf{Teleop})         & 4\%  & 8\%  \\
    VOSA            &8\% &4\% \\
    Auditory Sparse (\textbf{AS})   &12\% &16\% \\
    Auditory Rich (\textbf{AR})     &0\% &8\% \\
    Visual Sparse (\textbf{VS})  &24\% & \textbf{48\%} \\
    Visual Rich (\textbf{VR})    & \textbf{52\%} & 16\% \\
    \bottomrule
  \end{tabular}
\end{table}

To evaluate our hypotheses, we used statistical methods appropriate to our experimental design. For comparisons involving multiple interface conditions within the same set of participants, we used a one-way repeated-measures ANOVA to test for statistically significant differences across conditions. When ANOVA indicated a significant main effect, we conducted Tukey's Honest Significant Difference (HSD) post-hoc analysis to identify pairwise differences between treatment levels.
For hypothesis tests involving ordinal data, such as Likert-style scale responses, we used the Wilcoxon signed-rank test for paired comparisons. This allowed us to assess directional effects without assuming normality, while still leveraging the within-subject nature of the study.
For hypotheses evaluated using preference data, we analyzed preference frequencies using a chi-square goodness-of-fit test. This test enabled us to assess whether observed interface preferences differed significantly from chance expectations.
We used a significance level of $0.05$ for all comparisons.


\subsection{Feedback improves intent alignment and reduces corrective inputs}
The Teleop condition was excluded from this analysis because, without autonomous intent inference, alignment and corrective input metrics are not defined.
\textbf{H1} posits that providing feedback improves alignment between the robot’s inferred intent and the user’s intended goal. Alignment was measured as the proportion of time steps in which the robot’s predicted goal matched the user's intended task goal. Statistical tests revealed a significant effect of feedback condition in both tasks (shelving: \textit{F}(4, 96) $=$ 8.15, \textit{p} $<$ 0.001; sorting: \textit{F}(4, 96) $=$ 20.89, \textit{p} $<$ 0.001). Post-hoc comparisons showed that all feedback interfaces significantly improved alignment relative to the VOSA baseline in both tasks (\textit{p} $<$ 0.05), with no significant differences among feedback conditions. These results indicate that the presence of feedback, rather than its modality or richness, was sufficient to enhance intent alignment across tasks, supporting H1 (Fig.~\ref{fig:combined}a).

\textbf{H2} posits whether providing feedback reduces corrective user inputs. We define corrective inputs as the number of switches per trial in the robot’s predicted user intent. Because the shared autonomy controller and intent inference model were identical across all feedback conditions, and the environment remained fixed during each trial, differences in intent switching arise from how users adjust their joystick inputs in response to the interface rather than from changes in the inference logic or environment. Since the robot’s intent estimate is conditioned on joystick input, switches in its predicted user goal reflect instability in the inferred intent induced by user control adjustments. Fewer switches therefore indicate faster convergence toward a stable shared intent.
Statistical tests revealed a significant effect of interface condition in both the shelving task (\textit{F}(4, 96) $=$ 18.56, \textit{p} $<$ 0.001) and the sorting task (\textit{F}(4, 96) $=$ 9.49, \textit{p} $<$ 0.001). Post-hoc comparisons showed that all feedback interfaces resulted in significantly fewer intent changes than VOSA in both tasks (all \textit{p} $<$ 0.01). The largest reductions were observed for VR and VS, followed by the auditory conditions; however, no significant differences were found among feedback interfaces themselves. These results indicate that the presence of feedback was the dominant factor in reducing corrective intervention, supporting H2 (Fig.~\ref{fig:combined}b).

\subsection{Participants report greater understanding and trust with visual feedback than auditory feedback}
To evaluate \textbf{H3} and \textbf{H4}, we analyzed participants' post-study Likert responses comparing the two modalities---visual and auditory. Participants rated how well visual and auditory feedback helped them understand the robot's plan and how much they trusted the robot under each modality. Specifically, we analyzed the responses for the questions ``The visual (auditory) feedback helped me \textbf{understand} the robot’s plan'' and ``I \textbf{trusted} the robot more when it provided visual (auditory) feedback''.
A Wilcoxon signed-rank test revealed that participants rated the visual feedback as significantly more helpful for understanding the robot's plan than the auditory feedback (\textit{W} = 31.0, \textit{p} = 0.001). Likewise, participants reported significantly higher trust in the robot when it provided visual feedback compared to auditory feedback (\textit{W} = 28.0, \textit{p} = 0.001). These results show strong support for our hypotheses, indicating improved understanding and trust in visual feedback interfaces.

\subsection{Feedback does not uniformly increase perceived trust}
To evaluate \textbf{H5}, we compared trust ratings (Fig.~\ref{fig:radar-plot}) for each feedback modality against the VOSA baseline using paired Wilcoxon signed-rank tests. For the sorting task, Sparse Visual Feedback led to a significant increase in trust compared to baseline (M = 5.52 vs. 4.56; \textit{W} = 35.5, raw \textit{p} = 0.008, Holm-corrected \textit{p} = 0.033), whereas Sparse Audio Feedback, Rich Visual Feedback, and Rich Audio Feedback did not yield significant improvements after correction. In the shelving task, none of the feedback modalities produced statistically significant increases in trust relative to baseline after Holm–Bonferroni correction, although Sparse Visual Feedback again showed the largest positive shift in mean trust (M = 5.28 vs. 4.40). Overall, these results indicate that feedback does not uniformly increase user trust; rather, trust improvements depend on both task context and feedback design, with sparse visual feedback being most effective in the more cluttered and ambiguous sorting task.

\subsection{User interface preferences depend on task complexity, modality, and feedback presence}
To evaluate \textbf{H6} and  \textbf{H7}, 
we analyzed participants’ interface preferences collected at the end of the study (Table~\ref{tab:interface_preferences}). 

To evaluate \textbf{H6}, a chi-square goodness-of-fit test showed that user preference distributions deviated significantly from uniform in both the shelving task ($\chi^2$ = 27.56, $p < 0.001$) and the sorting task ($\chi^2$ = 19.40, $p = 0.0016$). In the shelving task, a majority of participants preferred the Visual Rich interface (52\%), followed by Visual Sparse (24\%), with substantially fewer selections for the auditory interfaces and the no-feedback baselines. In contrast, in the sorting task, participants preferred the Visual Sparse interface (48\%), whereas Visual Rich was selected less often (16\%).
These findings indicate that preferences for feedback richness are task-dependent, reflecting a trade-off between interpretability and cognitive load, and provide mixed support for H6. 

To evaluate \textbf{H7}, we collapsed the preferences into three categories: visual feedback (Visual Rich, Visual Sparse), auditory feedback (Audio Rich, Audio Sparse), and no-feedback baselines (VOSA, Teleop). 
Statistical tests indicate significant distributions across these categories for both shelving ($\chi^2$ = 20.48, $p < 0.001$) and sorting ($\chi^2$ = 11.12, $p = 0.003$).
In both tasks, visual feedback interfaces were most frequently preferred, followed by auditory interfaces, with no-feedback baselines least frequently preferred. 
These results provide support for H7, indicating a consistent user preference hierarchy favoring visual feedback over auditory feedback, and both over no-feedback conditions.

\begin{figure}[!t]
  \centering
  \includegraphics[width=\columnwidth]{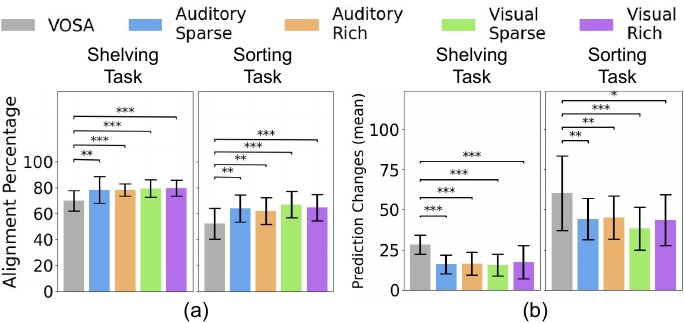}
  \caption{\textbf{Feedback effects on intent alignment and corrective behavior.} (a) Mean intent alignment percentage (higher is better), defined as the proportion of time steps in which the robot’s inferred goal matched the task-defined user goal. (b) Mean number of intent prediction changes per trial (lower is better). Error bars denote standard error; asterisks indicate significance levels (*$p<0.05$, **$p<0.01$, ***$p<0.001$).}
  \label{fig:combined}
\end{figure}

\section{Design Considerations for Transparent Shared Autonomy}
\textbf{Transparency design should emphasize goal legibility over belief disclosure.}
Across both tasks, the presence of feedback, regardless of modality or richness, consistently improved intent alignment and reduced corrective inputs relative to the no-feedback baseline (Fig.~\ref{fig:combined}). These results indicate that transparency enhances shared autonomy by stabilizing the inferred goal and reducing repeated user intervention. Notably, users did not require full access to the robot’s internal belief distribution to realize these coordination benefits; simply exposing the inferred goal was sufficient. This suggests that transparency need not be exhaustive to be effective.
Trust responses, however, were not uniformly related to the amount of information disclosed. Providing the full belief distribution did not consistently increase trust and, in some contexts, yielded no measurable gains. Instead, sparse visual feedback significantly increased trust in the more ambiguous sorting task, indicating that transparency supports trust most effectively when it clarifies uncertainty without overwhelming the user. These findings suggest that transparency does not automatically amplify trust in a shared autonomy system; rather, it supports trust by providing timely, interpretable, and task-aligned information.

\textbf{Favor visual feedback for continuous spatial interaction.}
While all feedback interfaces improved objective performance, visual feedback was consistently rated higher than auditory feedback in subjective measures of understanding and trust. Participants reported greater confidence in anticipating the robot’s actions and expressed higher trust when intent information was presented visually rather than verbally, indicating that feedback modality plays a critical role in how transparency is perceived.
This preference reflects the spatial and continuous nature of assistive manipulation tasks. Visual feedback allowed users to directly map the robot’s inferred intent onto the physical workspace and reference it persistently, enabling rapid interpretation without interrupting ongoing control. In contrast, auditory feedback, particularly when frequent, required users to process sequential information while simultaneously attending to joystick control and task execution, likely increasing cognitive load during fast-paced interaction. Several participants noted that this challenge was amplified when the robot’s predicted intent changed rapidly. One participant explained:
\textit{``I found the visual feedback to be more helpful. Auditory sparse was still okay, but the [auditory] rich was not helpful at all. It was easier to see the [intent] changes on the screen than to use spoken cues.''}
Together, these observations suggest that, in interaction settings where users continuously reference a visual workspace and display, visual feedback aligns naturally with ongoing perceptual processing.

\textbf{Match feedback richness to task complexity and environmental ambiguity.}
User preferences for sparse versus rich feedback were task-dependent. In the shelving task, participants overwhelmingly preferred the Visual Rich interface, whereas in the cluttered and ambiguous sorting task, participants favored the Visual Sparse interface.
This shift reveals an important trade-off between completeness and cognitive load. In simpler environments, rich feedback appeared to strengthen user trust by making the robot’s reasoning more transparent and predictable. However, in visually dense scenes with multiple objects, the same richness became counterproductive, overwhelming users with information. One participant explicitly commented on this contrast:
\textit{``In the [shelving] task, it was nice to see the confidence numbers as it made me understand the robot's thought process. But for the [sorting] task... 
I did not even look at the labels or the numbers.''} These findings highlight how users dynamically recalibrate their information needs in response to environmental complexity and reinforce the idea that transparency should be adaptive rather than static.

\textbf{Use auditory feedback selectively and sparingly.}  
Although auditory feedback was generally less preferred than visual feedback, it was not categorically ineffective. A subset of participants preferred sparse auditory feedback, reporting that it reduced the need to shift attention between the physical workspace and the visual display. For these users, brief spoken confirmations of intent changes allowed them to remain focused on the task without additional visual context switching. In such cases, sparse auditory cues were perceived as helpful, providing awareness of intent updates without introducing substantial cognitive burden.
However, the negative effect of auditory richness was particularly pronounced. Participants rarely selected the Auditory Rich interface as their preferred condition, and qualitative feedback indicated that verbose spoken explanations increased distraction rather than clarity—especially when the robot’s inferred intent was uncertain or changed frequently. One participant described this experience emphatically:
\textit{``I did not like the auditory rich feedback. It was so distracting that I couldn't concentrate on the task at all! I would not use this feedback if given a choice.''}
Unlike visual feedback, which users could reference selectively and at their own pace, auditory explanations unfolded continuously and competed directly with attentional resources during real-time control. When intent predictions fluctuated, repeated verbal updates further amplified this burden, disrupting control fluency rather than supporting understanding. These findings suggest that, during continuous shared autonomy interactions, auditory transparency may be most effective when limited to brief cues indicating salient intent changes rather than sustained verbal explanation.

\section{Conclusion and Future Work}
This work examined transparency as an interface design consideration in blended shared autonomy. By systematically varying feedback modality (visual vs.\ auditory) and information richness (sparse vs.\ rich) while holding the control policy fixed, we isolated how interface design shapes coordination, corrective intervention, and trust. Our results show that providing feedback improves intent alignment and reduces corrective input; exposing the inferred goal is sufficient to support efficient collaboration. However, revealing the robot’s full internal belief distribution does not consistently provide additional coordination or trust benefits.
Together, these findings clarify the functional role of transparency in shared autonomy: it improves coordination by stabilizing inferred intent and shapes user trust through task-appropriate information exposure rather than maximal disclosure. This provides concrete design guidance for future systems--transparency should prioritize goal legibility, and feedback modality and richness should be matched to task demands and user needs. Although evaluated in a vision-based system, these design principles extend to broader intent-inference-based shared autonomy settings and motivate adaptive transparency mechanisms that respond to context and user behavior during shared control.

Future work should investigate adaptive transparency strategies that dynamically adjust feedback modality and information richness based on uncertainty, task complexity, and user behavior. Longitudinal studies are needed to examine how transparency influences user trust and reliance over extended interaction. It is also important to explore modality trade-offs under degraded visual conditions, sensor misalignment, and alternative assistive contexts. Evaluating these principles in higher-dimensional manipulation tasks and with users with motor impairments will further clarify their applicability to real-world assistive robotics.

\section{Acknowledgments}
This work was conducted in the Aligned, Robust, and Interactive Autonomy (ARIA) Lab at the University of Utah. ARIA Lab research is supported in part by the NSF (IIS-2310759, IIS2416761), the NIH (R21EB035378), ARPA-H, and Coefficient Giving.
\addtolength{\textheight}{-2.5cm}   







\bibliographystyle{IEEEtran}
\bibliography{main}
\end{document}